\documentclass{article}
\usepackage{microtype}
\usepackage{graphicx}
\usepackage{subfigure}
\usepackage{booktabs}
\usepackage{amsmath,amssymb,mathtools}
\usepackage{hyperref}
\usepackage[capitalize,noabbrev]{cleveref}
\usepackage{caption}
\usepackage[preprint]{icml2026}

\begin{document}

\twocolumn[
\icmltitle{Interpreting V1 Population Activity via \\ Image--Neural Latent Representation Alignment}

\vskip 0.15in

\begin{center}

{\bfseries\normalsize
Xin Wang$^{1}$ \quad
Zhuangzhi Gao$^{2,3,4}$ \quad
Hongyi Qin$^{5,3}$\\[0.2em]
Zhongli Wu$^{3}$ \quad
Feixiang Zhou$^{\dagger,3,4}$ \quad
He Zhao$^{\dagger,3,4}$
}

\vskip 0.10in

{\small
$^{1}$School of Psychological and Cognitive Sciences, Peking University, Beijing, China\\
$^{2}$Department of Primary Care and Mental Health, University of Liverpool, Liverpool, United Kingdom\\
$^{3}$Liverpool Centre for Cardiovascular Science, University of Liverpool, Liverpool, United Kingdom\\
$^{4}$Department of Eye and Vision Sciences, University of Liverpool, Liverpool, United Kingdom\\
$^{5}$Institute of Life Course \& Medical Sciences, University of Liverpool, Liverpool, United Kingdom\\[0.2em]
$^{\dagger}$Corresponding authors: Feixiang Zhou \texttt{F.Zhou12@liverpool.ac.uk}; He Zhao \texttt{He.Zhao@liverpool.ac.uk}
}

\end{center}

\vskip 0.22in
]

\makeatletter
\begingroup
\long\def\@footnotetext#1{}
\long\def\footnotetext#1{}
\printAffiliationsAndNotice{}
\endgroup
\makeatother

\begin{abstract}
Understanding the neural mechanisms underlying visual computation has long been a central challenge in neuroscience.
Recent alignment-based approaches have improved the accuracy of decoding visual stimuli from brain activity, yet they provide limited insight into the neural computations that give rise to these improvements.
To address this gap, we propose Dual-Tower Image–Neural Alignment (DINA), an interpretable contrastive framework for analyzing population-level visual computations in primary visual cortex (V1).
DINA jointly trains a biologically motivated dual-tower architecture that aligns visual stimuli and corresponding V1 population responses in a shared latent space at the level of intermediate feature maps, enabling both accurate decoding and direct access to interpretable feature maps. 
Evaluated on large-scale two-photon calcium imaging data from mouse V1, DINA achieves accurate neural-based decoding while revealing that decoding performance is primarily supported by coarse, low-level visual structure, rather than semantic category information or fine-grained details.
Further analysis reveals that alignable feature maps emerge from multiple spatially distributed image regions, capturing both shape and texture cues, and are predominantly reconstructed by sparse subsets of strongly responsive neurons and their functional interactions. 
Together, these results confirm that, beyond enabling accurate decoding, DINA provides a principled framework for probing the computational mechanisms underlying visual processing in V1.
Code is available at \href{https://github.com/small-blingbling/DINA}{GitHub repository}.

\end{abstract}

\begin{figure}[t]
  \centering
  \includegraphics[width=\columnwidth]{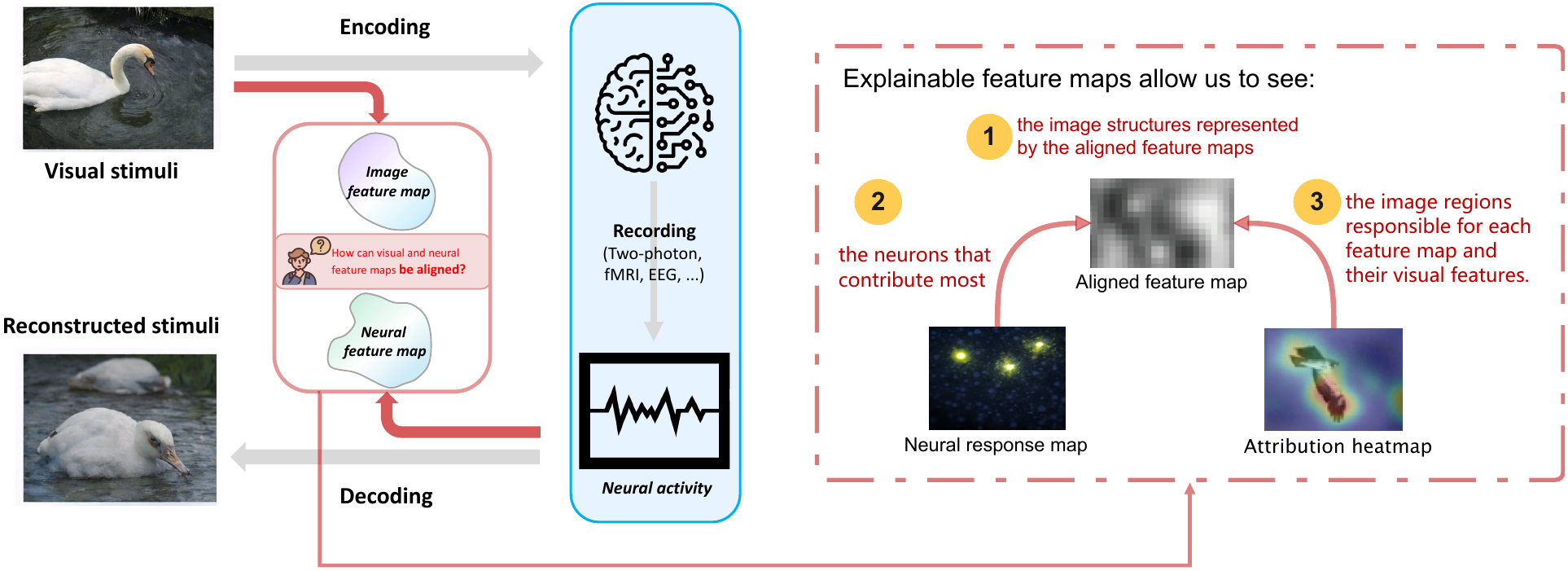}
  \caption
  {
   \textbf{Illustration of feature-level contrastive alignment.} 
Visual stimuli are transformed into intermediate image feature maps, while neural activities reconstruct intermediate neural feature maps.
Encoding and decoding can be viewed as aligning these heterogeneous representations in a shared feature space, raising the central question of how image-derived and neural-derived feature maps can be aligned.
Moreover, the resulting explainable feature maps enable interpretable analyses of population-level neural computation mechanisms.
}
  \label{fig:overview}
\end{figure}

\section{Introduction}

Understanding how visual information is processed by neural populations is a central goal of visual neuroscience. A key system for studying this problem is the primary visual cortex (V1), the first cortical stage of visual processing in the brain. V1 receives input from the retina via the thalamus and transforms visual signals into structured neural representations. Extensive prior work has established that V1 neurons exhibit selective tuning to basic visual features such as edges, orientations ~\cite{hubel1962receptive,hubel1968receptive}, and spatial frequency ~\cite{de1982spatial}. While these studies have provided important insights, they have largely focused on the tuning properties of individual neurons. Recent advances in recording technologies, particularly large-scale two-photon calcium imaging, now allow simultaneous measurements from large neural populations, shifting the study of neural computation from the single-neuron level to the population level.

At the population level, two complementary approaches are commonly used to study neural computation~\cite{kriegeskorte2019interpreting,naselaris2011encoding,mathis2024decoding}. Encoding approaches aim to predict neural responses from visual stimuli, providing insight into how visual information is represented in the brain. Decoding approaches, in contrast, attempt to reconstruct or identify visual stimuli from neural population activity, revealing what information can be read out from neural representations. While both paradigms have proven useful, decoding performance alone provides limited insight into the structure of the underlying representations or the computational strategies that give rise to them.

Existing encoding and decoding approaches typically model the relationship between visual stimuli and neural responses at the level of raw images~\cite{yamins2014performance,seeliger2018generative,xu2023multimodal,deng2024predicting}. However, neural responses do not directly represent visual images themselves; instead, they encode intermediate feature representations of the stimulus. This distinction is particularly important in early visual areas such as V1, where neural activities are shaped by local receptive fields and predominantly encode low-level visual structure. As illustrated in Fig.~1, both encoding and decoding can be viewed as operating on feature maps, rather than on raw images. 
\emph{Thus, a central question is how to align visual feature representations extracted from images with feature maps read out from neural population activity, such that corresponding image-neural pairs are close in latent space, while non-corresponding pairs are well separated.}
This feature-map-level alignment provides a natural entry point for probing population-level computational principles beyond decoding accuracy. More broadly, this perspective suggests that contrastive alignment should not be viewed merely as a means to improve decoding performance, but as a general analytical tool for dissecting population-level neural computation.

To operationalize this idea, we leverage large-scale two-photon calcium imaging recordings from mouse V1 and propose Dual-Tower Image–Neural Alignment (DINA), a framework that explicitly mirrors key computational motifs of the visual system. This biologically motivated design reflects the structure of cortical visual processing, such that feature maps extracted by the model are more likely to resemble those computed by the brain. The image tower incorporates parallel local and global processing pathways, and the neural tower models neuron-specific properties and population-level interactions. By aligning the latent feature maps produced by the two towers using a contrastive objective, the framework enables accurate decoding while obtaining intermediate representations suitable for probing population-level computational mechanisms in V1.

Using this framework, we show that successful image-neural alignment in V1 is primarily supported by coarse, low-dimensional visual structure rather than semantic category or high-frequency detail.
Through systematic and multi-level interpretability analyses of the aligned feature maps, we further find that image–neural alignment arises from multiple spatially distributed regions of the stimulus, within which diverse visual features, including both shape and texture cues, are jointly represented.
These feature maps are read out predominantly by sparse subsets of strongly responsive neurons and a small number of functionally important interactions.

\begin{figure*}[t]
    \centering
    \includegraphics[width=\textwidth]{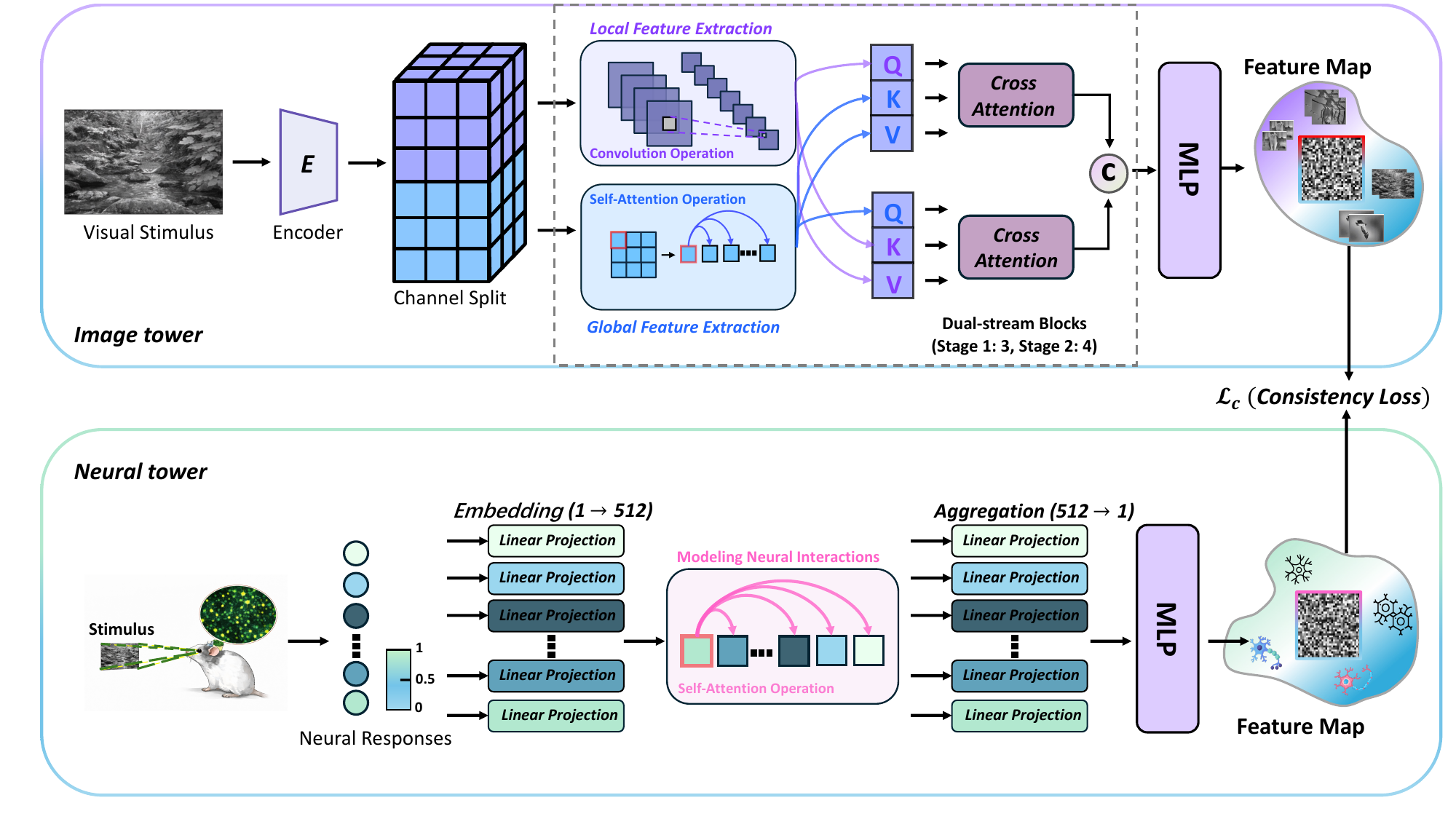}
    \caption{
    \textbf{Architecture of the DINA model.}
    An image tower and a neural tower independently project visual stimuli and V1 population responses into feature maps that are matched in dimensionality and aligned using a contrastive objective.
    }
    \label{fig:architecture}
\end{figure*}

\noindent \textbf{Our key contributions are as follows:}
\begin{itemize}
     \item{We formulate that contrastive image-neural alignment can be used not only for decoding, but as a general analytical tool for probing population-level neural computation.}

    \item{We propose a biologically grounded dual-tower image-neural model inspired by cortical visual processing, which achieves reliable image decoding while producing aligned feature maps in V1.}

    \item{Multi-level interpretability analyses of aligned image-neural feature maps reveal low-level visual structure, spatially distributed stimulus contributions, and sparse neural readout mechanisms in V1 population coding.}
\end{itemize}

\section{Related Work}
Prior work has largely focused on visual encoding, decoding, and  alignment-based approaches as separate directions, with an emphasis on predictive performance.
This section reviews these lines of work and shows how our approach bridges them through an interpretable alignment framework to probe population-level visual computation in primary visual cortex.

\subsection{Encoding and Decoding Models in Visual Cortex}

Early encoding models were based on hand-crafted feature representations, such as linear–nonlinear (LN) models~\cite{jones1987evaluation} and Gabor-based receptive field models~\cite{ringach2002spatial}, which captured basic tuning properties of neurons in early visual areas.
More recent work has predominantly relied on convolutional neural networks (CNNs) to predict neural responses in visual cortex~\cite{schrimpf2018brain,yamins2016using,anand2021quantifying,kindel2019learning,walker2019inception}. These models typically employ a fixed visual backbone, followed by a learned neural readout layer, and have achieved strong encoding performance at the population level.

Decoding approaches, in contrast, aim to infer the visual stimulus directly from neural population activity, framing neural responses as a code from which stimulus information can be read out~\cite{naselaris2011encoding}. 
Early decoding work primarily employed discriminative deep models to perform visual stimulus classification or identification from neural activity~\cite{kamitani2005decoding}.
More recent approaches have leveraged learned latent representations together with powerful generative priors to reconstruct complex natural images, 
including GAN-based models~\cite{guccluturk2017reconstructing,seeliger2018generative} 
and, more recently, diffusion-based models~\cite{takagi2023high,li2025neuraldiffuser}.

Although decoding accuracy provides an operational measure of the information present in neural activity, these approaches are typically optimized for task performance and offer limited insight into the structure of intermediate feature maps or the population-level computational principles that support successful decoding.

\subsection{Representation Alignment and Interpretability}

Recent work has increasingly adopted representation alignment frameworks to connect neural activity with visual stimuli, often using contrastive or multimodal embedding objectives.
These approaches have been successfully applied to neural-based image retrieval~\cite{rajabi2025human,scotti2024mindeye2}, stimulus identification~\cite{zhu2025fmri2ges,horikawa2017generic}, and reconstruction~\cite{ma2025brainclip}, demonstrating improved decoding performance.
However, in most existing work, alignment is primarily evaluated by decoding or retrieval accuracy, and the learned embedding space is treated as a black box optimized for performance. Moreover, the model architectures are typically designed for predictive accuracy rather than guided by principles of cortical visual processing, limiting the extent to which the resulting representations can be directly interpreted in terms of population-level neural computation.

In contrast to prior alignment approaches that primarily optimize performance, our work treats alignment as an analytic tool for studying neural computational mechanisms. We formulate image–neural correspondence at the level of intermediate feature maps and introduce a biologically inspired dual-tower architecture with parallel local and global processing pathways, reflecting key principles of cortical visual processing. Together, these design choices reveal interpretable, aligned feature-map representations and provide insight into how population-level computations in V1.

\section{Methods}

We introduce a dual-tower framework in which an image tower and a neural tower that aligns visual stimuli and neural population responses at the level of intermediate feature maps, as illustrated in Fig.~2.

The image tower adopts a dual-stream architecture to separately capture local visual structure and global contextual interactions.
Through integrated dual-stream blocks with inter-scale co-attention, the image tower produces compact intermediate feature maps.
In parallel, the neural tower embeds population responses by modeling neuron-specific properties and population-level interactions, and projects them into a latent feature map aligned with the image tower output.

The two towers are trained jointly using a contrastive objective that encourages alignment between matched image–neural trials and separation between mismatched trials.

\subsection{Image Tower}

To embed visual stimuli in a manner aligned with cortical visual processing, we adopt a dual-stream image encoder to separately model local fine-grained patterns and global long-range dependencies, following the structure of Dual-stream Network (DS-Net) with inter-scale co-attention ~\cite{mao2021dualstream}. This design is inspired by the parallel processing of local and global information in the visual system \cite{angelucci2002circuits,fink1997neural}.

In each dual-stream block, the input 
$\mathbf{F} \in \mathbb{R}^{C \times H \times W}$
is split along the channel dimension into a high-resolution local stream
$\mathbf{f}_l \in \mathbb{R}^{\frac{C}{2} \times H \times W}$
and a low-resolution global stream
$\mathbf{f}_g \in \mathbb{R}^{\frac{C}{2} \times H \times W}$,
which are processed in parallel via an intra-scale propagation module.
The local stream focuses on spatially localized structure, while the global stream captures long-range contextual interactions.
Here, $H = 68$, $W = 270$, and $C$ is set to 64 and 128 for the first and second stages, respectively.

\textbf{Local stream.} The high-resolution local stream
is processed using a $3 \times 3$ depth-wise convolution to extract local visual
patterns and obtain $\mathbf{f}_L$:
\begin{equation}
\mathbf{f}_L(i,j) = \sum_{m,n} \mathbf{W}(m,n) \odot \mathbf{f}_l(i+m, j+n),
\end{equation}
where $\mathbf{W}(m,n), (m,n) \in \{-1,0,1\}$ denote the learnable convolutional
kernels, $\mathbf{W}(m,n)$ and $\mathbf{f}_l(i,j)$ are both $\frac{C}{2}$-dimensional vectors, and $\odot$ is element-wise product. 

\textbf{Global stream.} The low-resolution global stream is flattened into a sequence of tokens and processed using self-attention:
\begin{equation}
\mathbf{Q}_g=\mathbf{f}_g\mathbf{W}_Q,\quad
\mathbf{K}_g=\mathbf{f}_g\mathbf{W}_K,\quad
\mathbf{V}_g=\mathbf{f}_g\mathbf{W}_V,
\end{equation}
\begin{equation}
\mathbf{f}_G=\mathrm{Softmax}\!\left(\frac{\mathbf{Q}_g\mathbf{K}_g^\top}{\sqrt{d}}\right)\mathbf{V}_g,
\end{equation}
capturing global dependencies across the visual field.

\textbf{Dual-stream co-attention.} To explicitly align local and global representations, DS-Net employs an co-attention mechanism, where the local stream attends to the global stream and vice versa. This module enables bidirectional information exchange between $\mathbf{f}_L$ and $\mathbf{f}_G$, producing scale-aligned features that integrate fine-grained structure with global context.

Each dual-stream block consists of an intra-scale propagation module, where local and global streams are processed in parallel, followed by an inter-scale co-attention module that enables bidirectional information exchange between the two streams. Multiple dual-stream blocks are stacked to form a stage, with spatial resolution
reduced between stages. Following this design, we use a lightweight configuration
with two stages, where the first stage contains 3 blocks and the second stage
contains 4 blocks. Given an input image of size $68\times270$, it is progressively
transformed into a feature map of size $128 \times 16 \times 64$ after the two
stages, and then further projected by an MLP into a single-channel feature map of
size $1 \times 16 \times 64$, which preserves the spatial structure of the feature maps
for interpretability. The resulting feature map is treated as the image latent representations for subsequent alignment with neural population representations.

\subsection{Neural Tower}

The neural tower aims to reconstruct a latent representation from population-level neural responses whose dimensionality matches that of the image tower. The input to the neural tower is the population response vector $\mathbf{r}\in\mathbb{R}^N$ corresponding to a single stimulus condition, where $N$ denotes the number of simultaneously recorded neurons. Specifically, the neural tower comprises neuron-wise feature extraction, inter-neuron interaction modeling, and neuron-wise aggregation to scalars, motivated by biological plausibility.

\textbf{Embedding.} 
The population response vector $\mathbf{r} \in \mathbb{R}^N$ is first mapped into a set of neuron embeddings. 
Each neuron $i$ is associated with an independent learnable embedding vector 
$\mathbf{w}_i^{\mathrm{emb}} \in \mathbb{R}^{d_{\mathrm{model}}}$, allowing the model to capture neuron-specific response properties. 
The embedding operation is defined as
\begin{equation}
\mathbf{e}_i = r_i \mathbf{w}_i^{\mathrm{emb}}, \qquad \mathbf{e}_i \in \mathbb{R}^{d_{\mathrm{model}}}.
\end{equation}

The embeddings of all neurons are then stacked to form the embedding matrix
\begin{equation}
\mathbf{R}^{\mathrm{emb}} =
\big[ \mathbf{e}_1, \mathbf{e}_2, \ldots, \mathbf{e}_N \big]^\top
\in \mathbb{R}^{N \times d_{\mathrm{model}}}.
\end{equation}
which serves as the input to subsequent population-level processing.

\textbf{Self-attention.}
Following the embedding stage, the neuron embedding matrix $\mathbf{R}^{emb} \in \mathbb{R}^{N \times d_{\mathrm{model}}}$ is processed using self-attention to model interactions within the neural population. Self-attention enables each neuron token to selectively attend to others, allowing informative responses to be integrated and amplified at the population level.

Specifically, query, key, and value projections are computed as
\begin{equation}
\mathbf{Q} = \mathbf{R}^{emb} \mathbf{W}_Q, \quad 
\mathbf{K} = \mathbf{R}^{emb} \mathbf{W}_K, \quad
\mathbf{V} = \mathbf{R}^{emb} \mathbf{W}_V,
\end{equation}
where $\mathbf{W}_Q, \mathbf{W}_K, \mathbf{W}_V \in \mathbb{R}^{d_{\mathrm{model}} \times d_{\mathrm{model}}}$ are learnable projection matrices.

The self-attention output is then given by
\begin{equation}
\mathbf{R}^{atten}= \mathrm{Softmax}\!\left( \frac{\mathbf{Q}\mathbf{K}^\top}{\sqrt{d_{\mathrm{model}}}} \right)\mathbf{V},
\end{equation}
producing an updated representation that captures pairwise interactions among neurons.

\textbf{Aggregation.}
Following the self-attention module, each neuron token encodes interaction-aware
information in a $d_{\mathrm{model}}$-dimensional space. In the next step, we
aggregate each neuron’s representation into a scalar value using a neuron-specific
aggregation layer.

Specifically, let $\mathbf{R}^{atten} \in \mathbb{R}^{N \times d_{\mathrm{model}}}$
denote the output of the self-attention module, where $\mathbf{r}_i^{atten}$
corresponds to the representation of the $i$-th neuron. Each neuron is associated
with an independent learnable aggregation vector
$\mathbf{w}_i^{\mathrm{agg}} \in \mathbb{R}^{d_{\mathrm{model}}}$, and the aggregated
response is computed as
\begin{equation}
u_i = \mathbf{w}_i^{\mathrm{agg}} \cdot \mathbf{r}_i^{atten}.
\end{equation}
The aggregated responses of all neurons are then stacked to form the output vector
\begin{equation}
\mathbf{R}^{agg} = [u_1, u_2, \ldots, u_N]^\top \in \mathbb{R}^N.
\end{equation}

\textbf{Multi-layer perceptron.}
The aggregated population response vector $\mathbf{R}^{agg} \in \mathbb{R}^N$ is projected by a multilayer perceptron into a neural latent vector, which is then reshaped into a $16 \times 64$ latent feature map. The resulting feature map matches the dimensionality of the image tower output and is used for subsequent alignment.

\begin{figure*}[t]
    \centering
    \includegraphics[width=\textwidth]{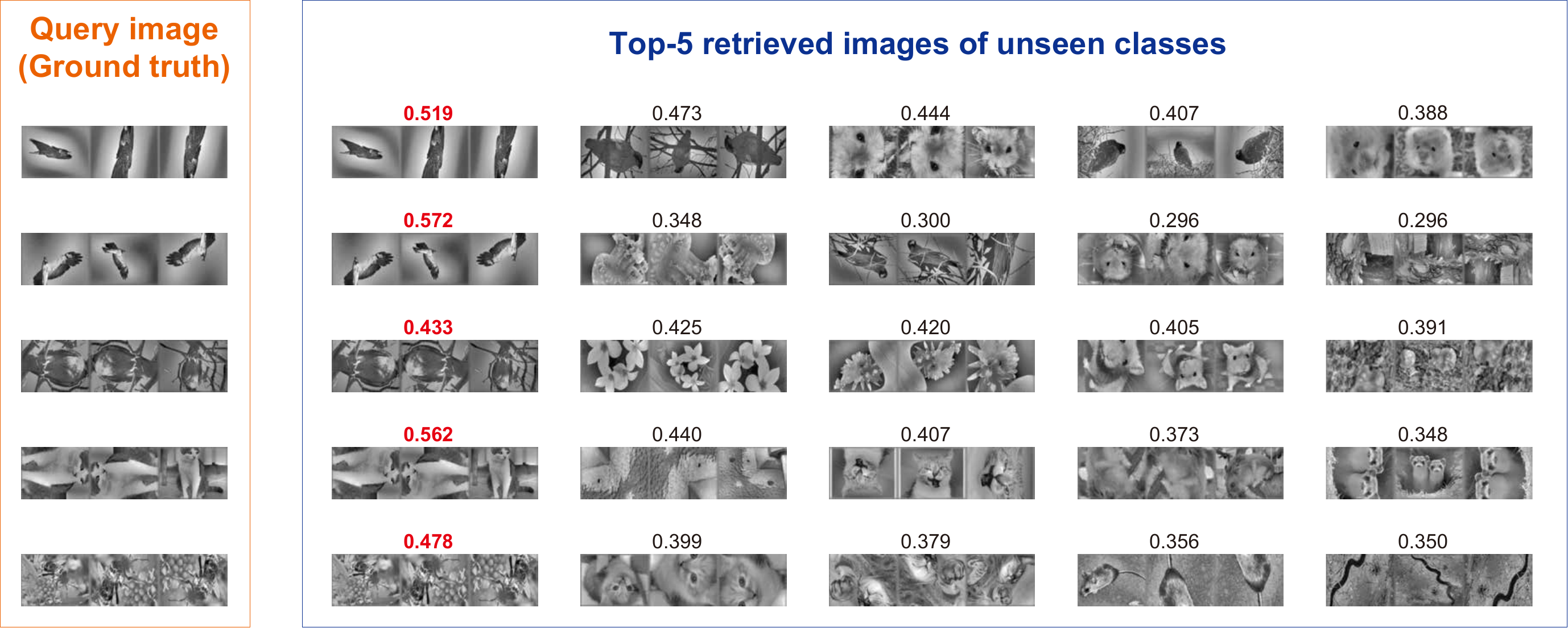}
    \caption{
     \textbf{Top-5 image retrieval results for representative neural queries from the test set.}
    }
    \label{fig:alignment}
\end{figure*}

\subsection{Dual-Tower Training}

We randomly split image-response pairs into training and validation sets, holding out 10\% of the data for test. Both towers are optimized using a contrastive alignment objective computed over batches of size 128 based on the InfoNCE loss with a fixed temperature of 0.01~\cite{radford2021learning}.
Optimization is performed using AdamW for the image tower and RMSprop for the neural tower. Models are trained for a fixed number of epochs with early stopping based on validation loss.

\begin{figure}[t]
    \centering
    \includegraphics[width=0.95\columnwidth]{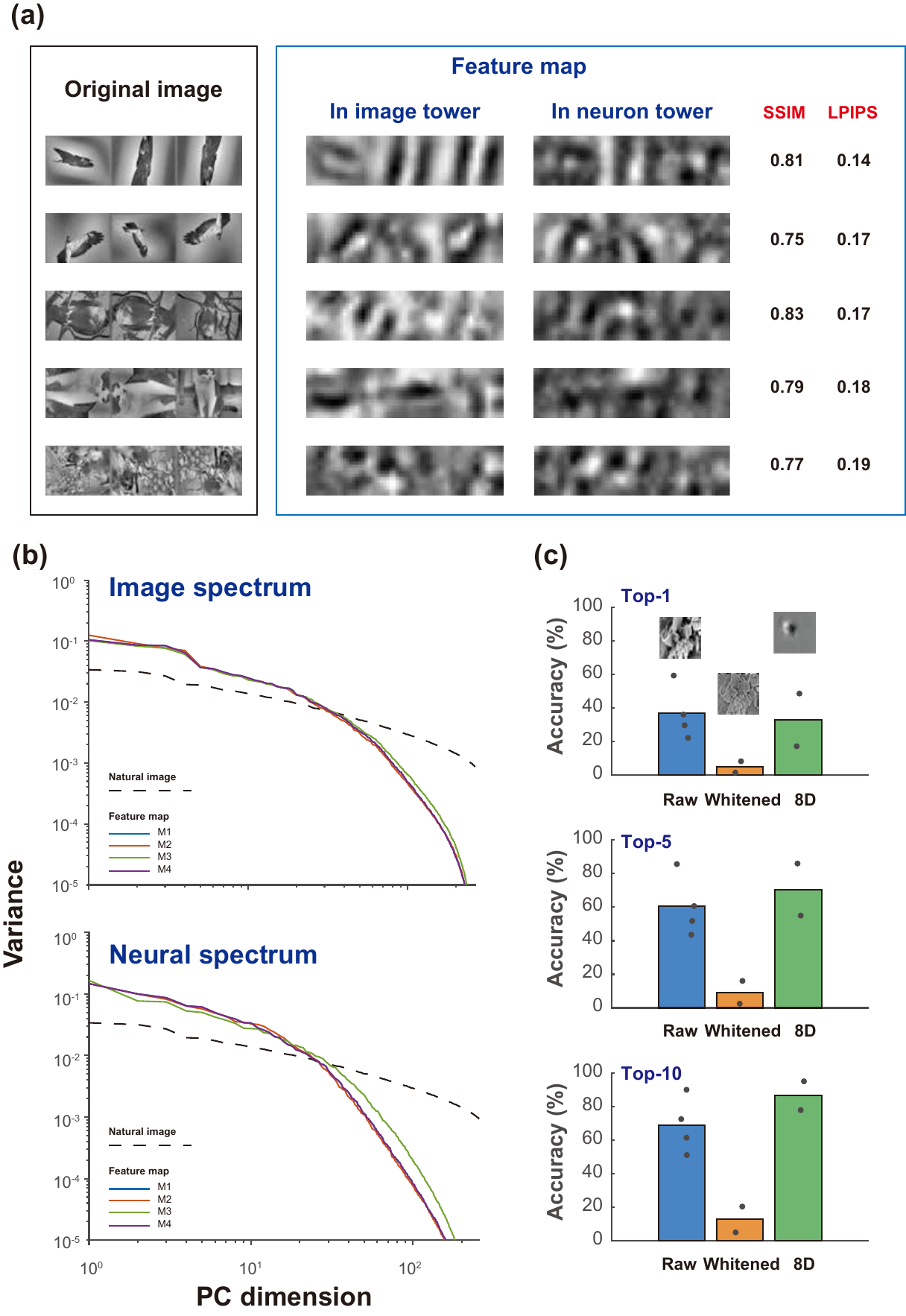}
    \caption{
    \textbf{Image-neural alignment reflects coarse, low-level visual structure.}
    (a) Latent feature maps produced by the image and neural towers for corresponding stimuli.
    (b) Variance spectra of natural images and aligned feature maps.
    (c) Image retrieval accuracy under different stimulus conditions, including natural, whitened, and low-dimensional (8D) images.
    }
    \label{fig:alignment}
\end{figure}

\begin{figure*}[t]
    \centering
    \includegraphics[width=\textwidth]{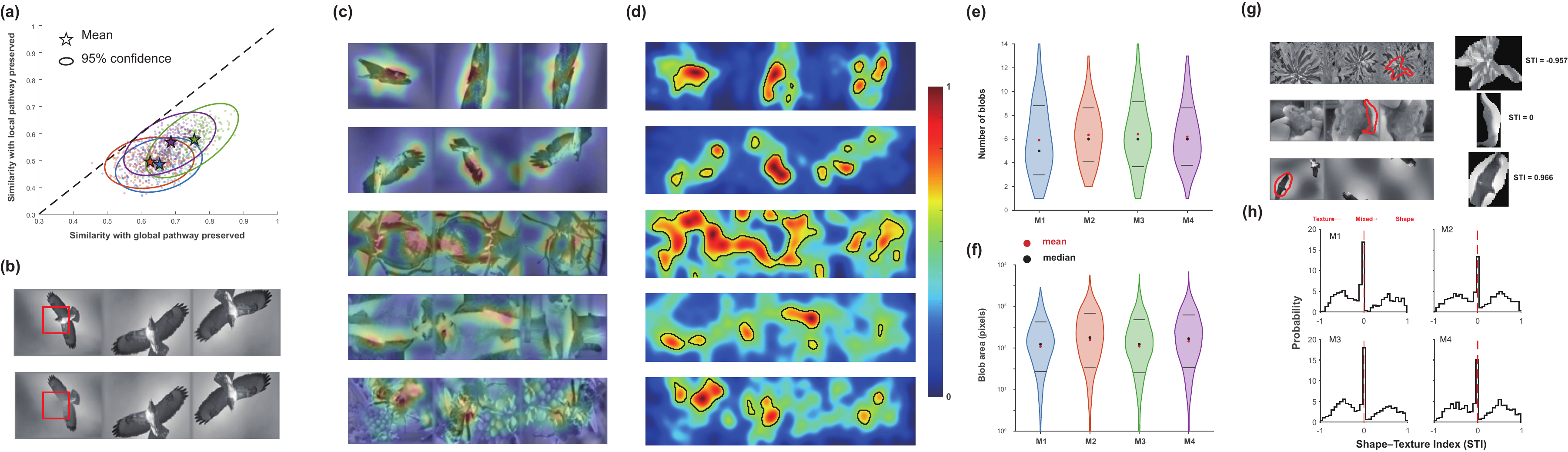}
    \caption{
     \textbf{Image-tower interpretability reveals distributed spatial contributions to aligned feature maps.}
    (a) Pathway masking analysis showing the structural similarity (SSIM) between feature maps obtained after masking either the local or global pathway and the original feature map.
    (b) Gaussian-windowed occlusion analysis applied to a representative stimulus image to estimate spatial contributions, using smooth kernels to avoid edge artifacts.
    (c) Occlusion-derived importance maps for individual stimuli, illustrating stimulus-dependent spatial contribution patterns.
    (d) Identification of informative regions (blobs) as spatially contiguous areas enclosed by the 0.6 isocontour of the importance map.
    (e) Distribution of the number of blobs per image across animals.
    (f) Distribution of blob sizes.
    (g) Representative examples of informative blobs with different Shape-Texture Index (STI) values.
    (h) Distributions of STI values across four mice (M1-M4); vertical dashed lines indicate STI = 0, separating texture- and shape-dominated regimes.
    }
    \label{fig:image_tower_interp}
\end{figure*}

\section{Experiments}

\subsection{Dataset}

We evaluated our method on publicly available two-photon calcium imaging recordings from mouse primary visual cortex (V1)~\cite{stringer2019high,stringer2018responses}, comprising population-level neural activities to a sequence of 2,800 image stimuli with approximately 10,000 neurons simultaneously recorded per animal. Stimuli were presented on three surrounding screens covering a wide visual field, resulting in visual images of size $68 \times 270$ pixels. Each image was presented at least twice per recording to improve the signal-to-noise ratio of trial-averaged neural responses.
Our experiments include recordings from four mice viewing natural images, two mice viewing whitened images, and two mice viewing low-dimensional (8D) images, allowing us to systematically assess alignment under different stimulus conditions.

Neural responses are provided as stimulus-averaged activity, resulting in a response matrix of size $(S \times N)$, where $S$ denotes the number of stimuli and $N$ the number of recorded neurons. The corresponding visual stimuli are represented as grayscale images defined over the mouse visual field, with dimensions of $68\times 270$.
Models are trained and evaluated independently for each mouse without cross-animal sharing.

\subsection{Evaluation}

Decoding accuracy was evaluated using an image retrieval conditioned on neural population responses.
During testing, each neural population response as a query was used to decode the corresponding visual stimulus by retrieving the most similar images in the embedding space from a set of 280 test images. Decoding was considered successful if the ground-truth image appeared among the top-$k$ retrieved results.
We report top-$1$, top-$5$, and top-$10$ decoding accuracy.

\subsection{Image-neural alignment}

We first evaluate image-neural alignment using a 280-way image retrieval. 
As summarized in Table~1, the DINA achieves a mean top-1 accuracy of 36.8\% and a mean top-5 accuracy of 60.5\% across four mice, demonstrating reliable decoding of visual stimuli from V1 population responses; corresponding image--neural similarity matrices are shown in Appendix Fig.~A.1.

We compare it with baseline models that employ standard visual encoders, including VGG-16 ~\cite{simonyan2015very} and ViT~\cite{dosovitskiy2021image}, paired with a feedforward neural readout implemented as MLP, a widely adopted choice for projecting neural signals to outputs ~\cite{stringer2021high}.
The training objective and overall optimization procedure are kept identical across models.

\begin{table*}[t]
\centering
\caption{Comparison of image retrieval accuracy (\%) using neural-derived representations across different structures.
Results are reported for four mice (M1-M4) and averaged across animals.
Top-$k$ accuracy is evaluated in a 280-way image retrieval task.}
\label{tab:retrieval_comparison}
\begin{tabular}{c|ccc|ccc|ccc}
\toprule
 & \multicolumn{9}{c}{\textbf{Dual-Tower Structure}} \\
\cmidrule(lr){2-10}
\textbf{Mouse} 
& \multicolumn{3}{c}{\textbf{Ours}} 
& \multicolumn{3}{c}{\textbf{VGG-16 - MLP}} 
& \multicolumn{3}{c}{\textbf{ViT - MLP}} \\
\cmidrule(lr){2-4} \cmidrule(lr){5-7} \cmidrule(lr){8-10}
 & Top-1 & Top-5 & Top-10 
 & Top-1 & Top-5 & Top-10
 & Top-1 & Top-5 & Top-10 \\
\midrule
M1 & 36.07 & 60.71 & 72.50 & 22.50 & 52.86 & 66.79 & 0.71 & 5.71 & 10.36 \\
M2 & 22.14 & 43.57 & 43.57 & 13.93 & 38.93 & 47.14 & 1.43 & 8.93 & 13.57 \\
M3 & 59.29 & 85.71 & 90.00 & 49.64 & 81.79 & 92.14 & 2.14 & 7.14 & 13.21 \\
M4 & 29.64 & 51.79 & 61.43 & 20.71 & 50.71 & 63.93 & 0.71 & 5.00 & 9.64 \\
\midrule
Mean & \textbf{36.79} & \textbf{60.45} & \textbf{68.75}
     & 26.70 & 56.57 & 67.50
     & 1.25 & 6.70 & 11.70 \\
\bottomrule
\end{tabular}
\end{table*}

Both the VGG-16–MLP and ViT–MLP baselines achieve lower retrieval accuracy than our approach across all four mice and top-$k$ settings, summarized in Table 1.
Among the baselines, the VGG-16–MLP model attains decoding accuracy that is only slightly lower than that of our model.
However, its aligned feature maps lack clear spatial organization and exhibit noise-like patterns, compared to those produced by our model (Appendix Fig.~A.2).

Qualitative retrieval results further illustrate the shared latent representations. 
The top-5 retrieved images for a given neural query often differ in semantic category while sharing similar visual structure (Fig.~3). 
To probe which visual features support accurate decoding, we examined the latent feature maps produced by the image and neural towers. Despite arising from distinct input modalities, the two feature maps exhibit highly consistent spatial organization (Fig.~4a). Quantitatively, across the four mice, aligned image- and neural-derived feature maps showed robust structural correspondence together with low perceptual distance. Structural SSIM (mean ± std) ranged from 0.655 ± 0.065 to 0.757 ± 0.068 across 4 mice, while LPIPS (mean ± std) ranged from 0.177 ± 0.023 to 0.229 ± 0.023, indicating strong alignment of visual structure.

To further characterize the shared structure, we analyzed the variance spectra of natural images and aligned feature maps using principal component analysis.
Relative to natural images, both image- and neural-derived feature maps show greater variance concentration in the leading principal components and a more rapid decay in higher-order components (Fig.~4b).
This spectral bias indicates that alignment between image and neural representations is dominated by coarse, low-level visual structure rather than fine-grained detail or semantic category information.

To test whether low-level visual structure plays a causal role in decoding performance, we assessed the dependence of retrieval accuracy on different level components using controlled stimulus manipulations (Fig.~4c).
Removing low-dimensional structure via image whitening leads to a sharp drop in retrieval accuracy from neural-derived representations, with mean top-1, top-5, and top-10 accuracies of 4.82\%, 9.29\%, and 12.68\%, respectively.
In contrast, retrieval performance is largely preserved for low-dimensional (8D) images that retain coarse structure while discarding fine detail, achieving mean top-1, top-5, and top-10 accuracies of 32.86\%, 70.54\%, and 86.43\%, respectively.
Both stimulus manipulations follow established protocols for controlling image statistics in studies of visual cortical representations~\cite{stringer2019high}.

These results support that accurate decoding of visual stimuli from V1 population activity is dominated by coarse, low-level visual structure rather than fine-grained detail, consistent with known response properties of V1 \cite{simoncelli2001natural,devalois1982spatial,hubel1962receptive} and supporting the biological plausibility of the aligned feature maps.

\subsection{Interpretability of Image Tower}

To understand which components of the image tower are critical for producing V1-aligned feature maps, we conduct a series of interpretability analyses focusing on pathway ablation and spatial contribution patterns.

We first assess the relative contribution of the local and global pathways by selectively masking one pathway at a time and measuring the structural similarity (SSIM) between the resulting feature map and the original one.
Across stimuli and mice, preserving the global pathway consistently results in substantially higher SSIM compared to preserving only the local pathway (Fig.~5a).
This indicates that the global pathway plays a dominant role in shaping the aligned feature maps.

To localize regions that contribute most strongly to the aligned feature maps, we perform a spatial occlusion analysis.
Local image regions are occluded using a Gaussian window that is slid across the entire image (Fig.~5b), and the resulting changes are quantified using structual SSIM. The Gaussian window is used to avoid introducing sharp boundary artifacts during occlusion, and detailed occlusion kernel and stride settings are provided in Appendix~A.1.
Each stimulus produces a distinct occlusion-induced importance map, demonstrating strong stimulus dependence (Fig.~5c).

By thresholding the occlusion-induced importance map at 0.6, we identify spatially contiguous regions of high contribution, referred to as blobs (Fig.~5d).
We then quantify the statistical properties of these blobs across all test stimuli and four mice.
Each image typically contains a small number of informative blobs (mean $\pm$ std: $6.21 \pm 0.23$ per image; Fig.~5e), indicating that aligned representations are driven by a small set of distributed spatial locations rather than uniform integration across the entire image.
Blob sizes are relatively consistent across animals (mean $\pm$ std: $148.4 \pm 29.5$ pixels; Fig.~5f), corresponding to localized spatial extents given the original image resolution (68 $\times$ 270 pixels).

These findings indicate that V1 encodes visual information via parallel, localized features that are aggregated into a feature map, rather than through long-range spatial integration. This is consistent with prior evidence for sparse population coding and localized receptive fields in V1~\cite{hubel1962receptive,stringer2021high}.

We characterize the visual features of individual informative blobs using a Shape–Texture Index (STI), which integrates edge orientation consistency and contour tortuosity to quantify the relative contributions of shape and texture (Appendix~A.2 and A.3).
Negative STI values indicate texture-dominated regions, whereas positive values indicate shape-dominated regions (see examples in Fig.~5g).

We compute the STI for all blobs extracted from the test set.
Across four mice, STI distributions are consistently centered near zero with broad spread (Fig.~5h), indicating that informative blobs include texture-, shape-, and mixed-dominated regions, with mixed representations being most common.

Although texture sensitivity is classically attributed to V2 rather than V1~\cite{freeman2013functional}, our STI results reveal the presence of texture-dominated regions in V1-aligned feature maps, indicating that texture-related information is present in V1 population activity.

\subsection{Interpretability of Neural Tower}

We perform neuron-level ablation analyses by ranking neurons according to response magnitude and reconstructing aligned feature maps from subsets corresponding to different population percentages.
Neural contribution is quantified by computing structural SSIM relative to the full-population reconstruction.
Across all four mice, retaining only 12.00 ± 2.31\% (mean ± SEM) of the most responsive neurons was sufficient to recover high similarity (SSIM $\geq$ 0.9) in the aligned latent representation.

\begin{figure}[t]
    \centering
    \includegraphics[width=\columnwidth]{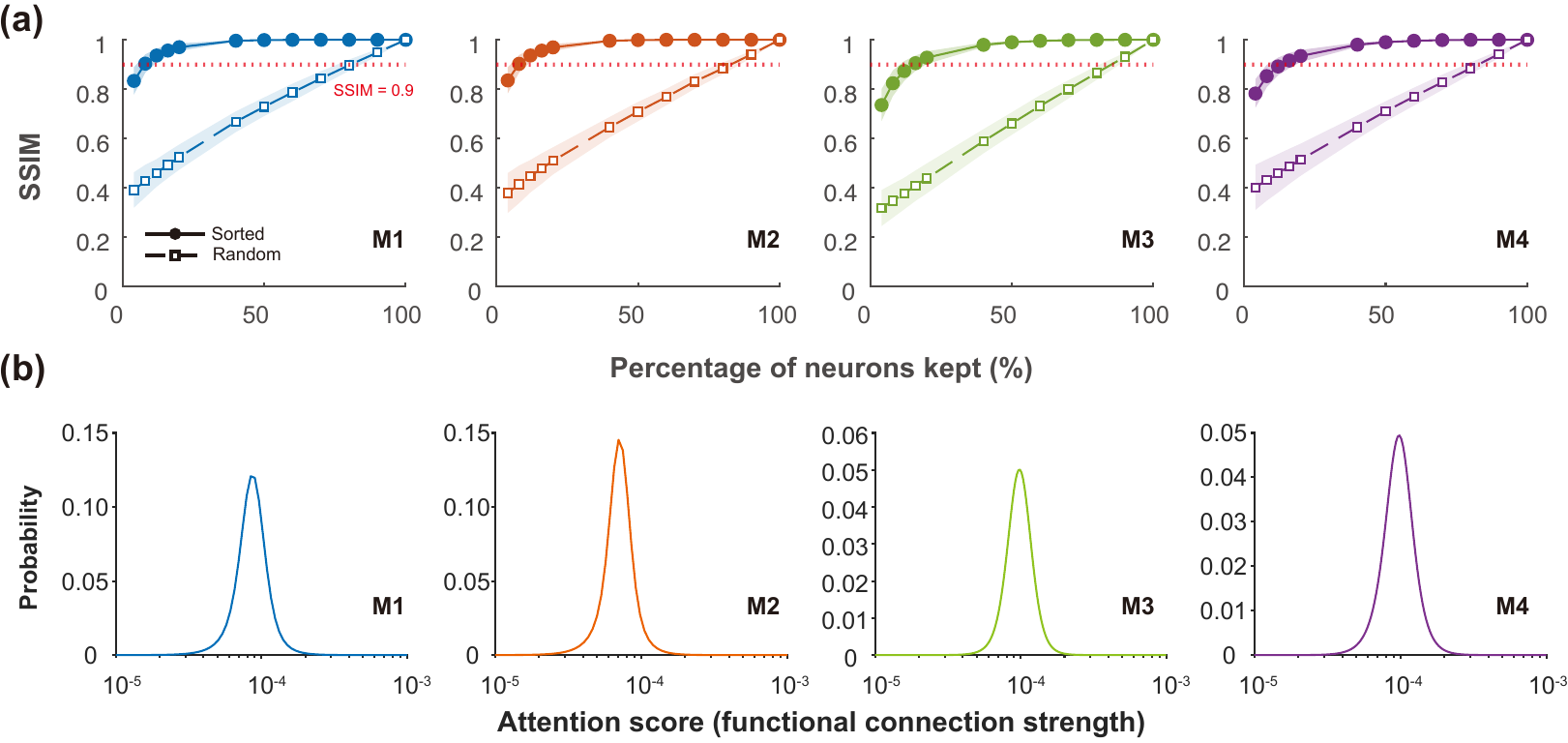}
    \caption{
     \textbf{Interpretability analysis of the neural tower.}
    (a) Neuron-level ablation analysis.
    Neurons are selected either by response magnitude (Sorted) or at random (Random), and subsets of increasing size are used to reconstruct the latent neural feature map.
    For each mouse (M1-M4), SSIM is computed between feature maps reconstructed from neuron subsets and from the full population.
    (b) Distributions of attention scores learned in the neural tower, interpreted as functional interaction strengths between neurons.
    }
    \label{fig:neuron_interp}
\end{figure}

As a control, we repeat the same analysis using randomly selected neuron subsets, averaging results over 10 independent samplings at each subset size.
In this case, substantially larger fractions of neurons are required to achieve comparable reconstruction similarity, and the recovery of feature map similarity is markedly slower than that observed for response-ranked subsets (Fig.~6a).
This contrast suggests that alignment is not supported by uniformly distributed population activity, but is instead driven by a small subset of highly responsive neurons.

Beyond identifying individual neuron contributions, we further examine interactions between neurons (functional connections) by analyzing the attention scores in the neural tower.
Across all four mice, the distribution of attention scores is highly skewed, closely resembling a log-normal distribution (Fig.~6b).
Most neuron pairs exhibit weak interactions, while a small fraction of connections carry disproportionately large weights.

The highly skewed, approximately log-normal distribution of attention weights provide convergent population-level evidence for prior hypotheses that cortical processing is sparse and efficient, with a small fraction of neurons and connections carrying most of the functional load~\cite{buzsaki2014lognormal,buzsaki2004neuronal}.

\section{Conclusion}

We introduced an interpretable contrastive framework that aligns image and neural representations at the level of intermediate feature maps to study population-level visual mechanisms in mouse V1.
The aligned intermediate feature maps are dominated by coarse, low-dimensional visual structure rather than semantic category or fine-grained detail, and integrate both shape and texture cues with contributions from multiple spatially distributed image regions.
Neural feature maps are read out by a sparse subset of highly responsive neurons with strongly non-uniform interactions, consistent with distributed coding in V1.
More broadly, this contrastive alignment framework is not limited to V1 and can be extended to compare latent feature maps across different brain regions, offering a pathway to probe hierarchical organization in the brain.

\section*{Imapact Statement}
Understanding population-level neural representations of visual stimuli has progressed steadily in recent years. However, models relating visual inputs to neural activity remain primarily scientific tools used in controlled research settings. The framework proposed in this work is intended to support mechanistic investigation of visual processing in primary visual cortex, not direct deployment or real-world decision-making. Accordingly, its outputs should be interpreted as descriptive analyses rather than actionable predictions.

This research may provide useful tools for studying sensory coding and population-level computation in neuroscience and may inform the development of more interpretable and biologically grounded machine learning models. Although such advances could eventually contribute to neural decoding or brain–computer interface research, the present work focuses on offline analysis and foundational scientific understanding.

\bibliography{references}

@article{stringer2019high,
  title={High-dimensional geometry of population responses in visual cortex},
  author={Stringer, Carsen and Pachitariu, Marius and Steinmetz, Nicholas and Carandini, Matteo and Harris, Kenneth D},
  journal={Nature},
  volume={571},
  number={7765},
  pages={361--365},
  year={2019},
  publisher={Nature Publishing Group UK London}
}

@inproceedings{mao2021dualstream,
  title={Dual-stream Network for Visual Recognition},
  author={Mao, Mingyuan and Gao, Peng and Zhang, Renrui and Zheng, Honghui and Ma, Teli and Peng, Yan and Ding, Errui and Zhang, Baochang and Han, Shumin},
  booktitle={Advances in Neural Information Processing Systems},
  year={2021}
}

@article{angelucci2002circuits,
  title={Circuits for local and global signal integration in primary visual cortex},
  author={Angelucci, Alessandra and Levitt, Jonathan B and Walton, Emma JS and Hupe, Jean-Michel and Bullier, Jean and Lund, Jennifer S},
  journal={Journal of Neuroscience},
  volume={22},
  number={19},
  pages={8633--8646},
  year={2002},
  publisher={Society for Neuroscience}
}

@article{fink1997neural,
  title={Neural mechanisms involved in the processing of global and local aspects of hierarchically organized visual stimuli.},
  author={Fink, Gereon R and Halligan, Peter W and Marshall, John C and Frith, Chris D and Frackowiak, RS and Dolan, Raymond J},
  journal={Brain: a journal of neurology},
  volume={120},
  number={10},
  pages={1779--1791},
  year={1997}
}

@article{hubel1962receptive,
  title={Receptive fields, binocular interaction and functional architecture in the cat's visual cortex},
  author={Hubel, David H and Wiesel, Torsten N},
  journal={The Journal of Physiology},
  year={1962}
}

@article{devalois1982spatial,
  title={Spatial frequency selectivity of cells in macaque visual cortex},
  author={De Valois, Russell L and Albrecht, Duane G and Thorell, Lewis G},
  journal={Vision Research},
  year={1982}
}

@article{simoncelli2001natural,
  title={Natural image statistics and neural representation},
  author={Simoncelli, Eero P and Olshausen, Bruno A},
  journal={Annual Review of Neuroscience},
  year={2001}
}

@inproceedings{rajabi2025human,
  title={Human-Aligned Image Models Improve Visual Decoding from the Brain},
  author={Rajabi, Nona and Ribeiro, Ant{\^o}nio H and Vasco, Miguel and Taleb, Farzaneh and Bj{\"o}rkman, M{\aa}rten and Kragic, Danica},
  booktitle={Forty-second International Conference on Machine Learning},
  year={2025}
}

@article{jones1987evaluation,
  title={An evaluation of the two-dimensional Gabor filter model of simple receptive fields in cat striate cortex},
  author={Jones, James P. and Palmer, Linda A.},
  journal={Journal of Neurophysiology},
  volume={58},
  number={6},
  pages={1233--1258},
  year={1987}
}

@article{yamins2014performance,
  title={Performance-optimized hierarchical models predict neural responses in higher visual cortex},
  author={Yamins, Daniel L. K. and Hong, Ha and Cadieu, Charles F. and Solomon, Ethan A. and Seibert, Darren and DiCarlo, James J.},
  journal={Proceedings of the National Academy of Sciences},
  volume={111},
  number={23},
  pages={8619--8624},
  year={2014}
}

@article{kindel2019learning,
  title={Learning receptive fields of individual neurons in visual cortex by deep learning},
  author={Kindel, William F. and Christensen, Eric D. and Zylberberg, Joel},
  journal={PLoS Computational Biology},
  volume={15},
  number={6},
  pages={e1006897},
  year={2019}
}

@article{walker2019inception,
  title={Inception loops discover what excites neurons most using deep predictive models},
  author={Walker, Edgar Y. and Sinz, Fabian H. and Cobos, Esteban and Muhammad, Talia and Froudarakis, Emmanouil and Miller, Kathleen D. and Tolias, Andreas S.},
  journal={Journal of Neuroscience},
  volume={39},
  number={32},
  pages={6418--6432},
  year={2019}
}

@article{de1982spatial,
  title={Spatial frequency selectivity of cells in macaque visual cortex},
  author={De Valois, Russell L and Albrecht, Duane G and Thorell, Lisa G},
  journal={Vision research},
  volume={22},
  number={5},
  pages={545--559},
  year={1982},
  publisher={Elsevier}
}

@article{kamitani2005decoding,
  title={Decoding the visual and subjective contents of the human brain},
  author={Kamitani, Yukiyasu and Tong, Frank},
  journal={Nature neuroscience},
  volume={8},
  number={5},
  pages={679--685},
  year={2005},
  publisher={Nature Publishing Group US New York}
}

@inproceedings{takagi2023high,
  title={High-resolution image reconstruction with latent diffusion models from human brain activity},
  author={Takagi, Yu and Nishimoto, Shinji},
  booktitle={Proceedings of the IEEE/CVF conference on computer vision and pattern recognition},
  pages={14453--14463},
  year={2023}
}

@article{horikawa2017generic,
  title={Generic decoding of seen and imagined objects using hierarchical visual features},
  author={Horikawa, Tomoyasu and Kamitani, Yukiyasu},
  journal={Nature communications},
  volume={8},
  number={1},
  pages={15037},
  year={2017},
  publisher={Nature Publishing Group UK London}
}

@inproceedings{radford2021learning,
  title={Learning transferable visual models from natural language supervision},
  author={Radford, Alec and Kim, Jong Wook and Hallacy, Chris and Ramesh, Aditya and Goh, Gabriel and Agarwal, Sandhini and Sastry, Girish and Askell, Amanda and Mishkin, Pamela and Clark, Jack and others},
  booktitle={International conference on machine learning},
  pages={8748--8763},
  year={2021},
  organization={PmLR}
}

@dataset{stringer2018responses,
  title        = {Responses of ten thousand neurons to 2,800 natural images},
  author       = {Stringer, Carsen and Pachitariu, Marius and Carandini, Matteo and Harris, Kenneth D.},
  year         = {2018},
  publisher    = {Figshare},
  doi          = {10.25378/janelia.6845348.v3},
  url          = {https://doi.org/10.25378/janelia.6845348.v3}
}

@inproceedings{simonyan2015very,
  title={Very deep convolutional networks for large-scale image recognition},
  author={Simonyan, Karen and Zisserman, Andrew},
  booktitle={International Conference on Learning Representations (ICLR)},
  year={2015}
}

@inproceedings{dosovitskiy2021image,
  title={An image is worth 16x16 words: Transformers for image recognition at scale},
  author={Dosovitskiy, Alexey and Beyer, Lucas and Kolesnikov, Alexander and Weissenborn, Dirk and Zhai, Xiaohua and Unterthiner, Thomas and Dehghani, Mostafa and Minderer, Matthias and Heigold, Georg and Gelly, Sylvain and Uszkoreit, Jakob and Houlsby, Neil},
  booktitle={International Conference on Learning Representations (ICLR)},
  year={2021}
}

@article{hubel1968receptive,
  title={Receptive fields and functional architecture of monkey striate cortex},
  author={Hubel, David H and Wiesel, Torsten N},
  journal={The Journal of physiology},
  volume={195},
  number={1},
  pages={215--243},
  year={1968},
  publisher={Wiley Online Library}
}

@article{ringach2002spatial,
  title={Spatial structure and symmetry of simple-cell receptive fields in macaque primary visual cortex},
  author={Ringach, Dario L},
  journal={Journal of neurophysiology},
  volume={88},
  number={1},
  pages={455--463},
  year={2002},
  publisher={American Physiological Society Bethesda, MD}
}

@article{yamins2016using,
  title={Using goal-driven deep learning models to understand sensory cortex},
  author={Yamins, Daniel LK and DiCarlo, James J},
  journal={Nature neuroscience},
  volume={19},
  number={3},
  pages={356--365},
  year={2016},
  publisher={Nature Publishing Group}
}

@article{schrimpf2018brain,
  title={Brain-score: Which artificial neural network for object recognition is most brain-like?},
  author={Schrimpf, Martin and Kubilius, Jonas and Hong, Ha and Majaj, Najib J and Rajalingham, Rishi and Issa, Elias B and Kar, Kohitij and Bashivan, Pouya and Prescott-Roy, Jonathan and Geiger, Franziska and others},
  journal={BioRxiv},
  pages={407007},
  year={2018},
  publisher={Cold Spring Harbor Laboratory}
}

@article{anand2021quantifying,
  title={Quantifying the brain predictivity of artificial neural networks with nonlinear response mapping},
  author={Anand, Aditi and Sen, Sanchari and Roy, Kaushik},
  journal={Frontiers in Computational Neuroscience},
  volume={15},
  pages={609721},
  year={2021},
  publisher={Frontiers Media SA}
}

@article{ma2025brainclip,
  title={BrainCLIP: Brain representation via CLIP for generic natural visual stimulus decoding},
  author={Ma, Yongqiang and Liu, Yulong and Chen, Liangjun and Zhu, Guibo and Chen, Badong and Zheng, Nanning},
  journal={IEEE Transactions on Medical Imaging},
  year={2025},
  publisher={IEEE}
}

@article{zhu2025fmri2ges,
  title={fMRI2GES: Co-speech Gesture Reconstruction from fMRI Signal with Dual Brain Decoding Alignment},
  author={Zhu, Chunzheng and Shao, Jialin and Lin, Jianxin and Wang, Yijun and Wang, Jing and Tang, Jinhui and Li, Kenli},
  journal={IEEE Transactions on Circuits and Systems for Video Technology},
  year={2025},
  publisher={IEEE}
}

@article{scotti2024mindeye2,
  title={Mindeye2: Shared-subject models enable fmri-to-image with 1 hour of data},
  author={Scotti, Paul S and Tripathy, Mihir and Villanueva, Cesar Kadir Torrico and Kneeland, Reese and Chen, Tong and Narang, Ashutosh and Santhirasegaran, Charan and Xu, Jonathan and Naselaris, Thomas and Norman, Kenneth A and others},
  journal={arXiv preprint arXiv:2403.11207},
  year={2024}
}

@article{naselaris2011encoding,
  title={Encoding and decoding in fMRI},
  author={Naselaris, Thomas and Kay, Kendrick N and Nishimoto, Shinji and Gallant, Jack L},
  journal={Neuroimage},
  volume={56},
  number={2},
  pages={400--410},
  year={2011},
  publisher={Elsevier}
}

@article{guccluturk2017reconstructing,
  title={Reconstructing perceived faces from brain activations with deep adversarial neural decoding},
  author={G{\"u}{\c{c}}l{\"u}t{\"u}rk, Ya{\u{g}}mur and G{\"u}{\c{c}}l{\"u}, Umut and Seeliger, Katja and Bosch, Sander and van Lier, Rob and van Gerven, Marcel A},
  journal={Advances in neural information processing systems},
  volume={30},
  year={2017}
}

@article{seeliger2018generative,
  title={Generative adversarial networks for reconstructing natural images from brain activity},
  author={Seeliger, Katja and G{\"u}{\c{c}}l{\"u}, Umut and Ambrogioni, Luca and G{\"u}{\c{c}}l{\"u}t{\"u}rk, Yagmur and Van Gerven, Marcel AJ},
  journal={NeuroImage},
  volume={181},
  pages={775--785},
  year={2018},
  publisher={Elsevier}
}

@article{li2025neuraldiffuser,
  title={NeuralDiffuser: Neuroscience-Inspired Diffusion Guidance for fMRI Visual Reconstruction},
  author={Li, Haoyu and Wu, Hao and Chen, Badong},
  journal={IEEE Transactions on Image Processing},
  year={2025},
  publisher={IEEE}
}

@article{kriegeskorte2019interpreting,
  title={Interpreting encoding and decoding models},
  author={Kriegeskorte, Nikolaus and Douglas, Pamela K},
  journal={Current opinion in neurobiology},
  volume={55},
  pages={167--179},
  year={2019},
  publisher={Elsevier}
}

@article{mathis2024decoding,
  title={Decoding the brain: From neural representations to mechanistic models},
  author={Mathis, Mackenzie Weygandt and Rotondo, Adriana Perez and Chang, Edward F and Tolias, Andreas S and Mathis, Alexander},
  journal={Cell},
  volume={187},
  number={21},
  pages={5814--5832},
  year={2024},
  publisher={Elsevier}
}

@article{xu2023multimodal,
  title={Multimodal deep learning model unveils behavioral dynamics of V1 activity in freely moving mice},
  author={Xu, Aiwen and Hou, Yuchen and Niell, Cristopher and Beyeler, Michael},
  journal={Advances in neural information processing systems},
  volume={36},
  pages={15341--15357},
  year={2023}
}

@article{deng2024predicting,
  title={Predicting Single Neuron Responses of the Primary Visual Cortex with Deep Learning Model},
  author={Deng, Kaiwen and Schwendeman, Peter S and Guan, Yuanfang},
  journal={Advanced Science},
  volume={11},
  number={15},
  pages={2305626},
  year={2024},
  publisher={Wiley Online Library}
}

@article{stringer2021high,
  title={High-precision coding in visual cortex},
  author={Stringer, Carsen and Michaelos, Michalis and Tsyboulski, Dmitri and Lindo, Sarah E and Pachitariu, Marius},
  journal={Cell},
  volume={184},
  number={10},
  pages={2767--2778},
  year={2021},
  publisher={Elsevier}
}

@article{freeman2013functional,
  title={A functional and perceptual signature of the second visual area in primates},
  author={Freeman, Jeremy and Ziemba, Corey M and Heeger, David J and Simoncelli, Eero P and Movshon, J Anthony},
  journal={Nature neuroscience},
  volume={16},
  number={7},
  pages={974--981},
  year={2013},
  publisher={Nature Publishing Group US New York}
}

@article{buzsaki2014lognormal,
  title   = {The log-dynamic brain: how skewed distributions affect network operations},
  author  = {Buzs{\'a}ki, Gy{\"o}rgy and Mizuseki, Kenji},
  journal = {Nature Reviews Neuroscience},
  volume  = {15},
  number  = {4},
  pages   = {264--278},
  year    = {2014}
}

@article{buzsaki2004neuronal,
  title   = {Neuronal oscillations in cortical networks},
  author  = {Buzs{\'a}ki, Gy{\"o}rgy},
  journal = {Science},
  volume  = {304},
  number  = {5679},
  pages   = {1926--1929},
  year    = {2004}
}
\bibliographystyle{icml2025}

\appendix
\renewcommand{\thefigure}{A.\arabic{figure}}
\setcounter{figure}{0}

\onecolumn
\section{Supplementary Computational Methods}
\subsection{Gaussian Occlusion-based Receptive Field Estimation}

Given an input image $I \in \mathbb{R}^{H \times W}$, we estimate a spatial receptive field (RF) map by measuring the sensitivity of the image tower features to localized Gaussian occlusions.

\paragraph{Gaussian-weighted occlusion.}
For each spatial location $(x_0, y_0)$ sampled on a regular grid with stride $s=2$, we define a two-dimensional isotropic Gaussian mask
\begin{equation}
G(x,y \mid x_0,y_0,\sigma)
=
\exp\!\left(
-\frac{(x-x_0)^2 + (y-y_0)^2}{2\sigma^2}
\right),
\end{equation}
where $\sigma = 4$ pixels.
The Gaussian is truncated to a window of size $6\sigma+1$ and peak-normalized such that $G(x_0,y_0)=1$.

The occluded image $I_{\text{occ}}$ is constructed by interpolating between the original image and the local mean intensity:
\begin{equation}
I_{\text{occ}}(x,y)
=
\left(1-\alpha G(x,y)\right) I(x,y)
+
\alpha G(x,y)\,\bar{I}_{\Omega},
\end{equation}
where $\alpha = 1.0$ controls the occlusion strength and $\bar{I}_{\Omega}$ denotes the mean intensity within the Gaussian support region $\Omega$.
This smooth Gaussian interpolation avoids introducing sharp artificial edges, which could otherwise confound feature perturbation measurements.

\paragraph{Feature perturbation measurement.}
Let $f(I)$ denote the feature representation produced by the image tower.
For each occluded image, we compute two complementary feature perturbation measures:
\begin{equation}
\Delta_{\text{cos}}
=
1 - \cos\!\left(
f(I), f(I_{\text{occ}})
\right),
\end{equation}
and
\begin{equation}
\Delta_{\text{ssim}}
=
1 - \operatorname{SSIM}\!\left(
f(I), f(I_{\text{occ}})
\right),
\end{equation}
where SSIM is evaluated using a $3\times3$ window and averaged across spatial locations.

\paragraph{Receptive field accumulation.}
For each occlusion center, the perturbation response is softly assigned to the spatial domain using the Gaussian mask:
\begin{equation}
R(x,y)
=
\frac{
\sum\limits_k \Delta_k \, G_k(x,y)
}{
\sum\limits_k G_k(x,y) + \varepsilon
},
\end{equation}
where $k$ indexes occlusion centers and $\Delta_k$ denotes either $\Delta_{\text{cos}}$ or $\Delta_{\text{ssim}}$.

The RF map is min-max normalized across the spatial domain:
\begin{equation}
\hat{R}(x,y)
=
\frac{
R(x,y) - \min(R)
}{
\max(R) - \min(R)
}.
\end{equation}

The occlusion procedure is repeated $N=5$ times with identical parameters, and the resulting RF maps are averaged to improve stability.

\subsection{Blob Extraction and Shape Analysis}

Salient RF regions are extracted by thresholding the normalized RF map:
\begin{equation}
\mathcal{B} = \{(x,y) \mid \hat{R}(x,y) > \tau \},
\end{equation}
where $\tau = 0.6$.
Connected components in $\mathcal{B}$ are identified as individual RF blobs.

Edge responses are computed within each blob using the Sobel operator, followed by binarization and skeletonization.
From the resulting skeleton, we extract quantitative shape descriptors including total edge length, fragmentation, average tortuosity, and orientation coherence.

\paragraph{Orientation coherence.}
For each RF blob, we compute the local edge orientation from the skeletonized edge map.
Given a connected skeleton segment with endpoints $(x_s,y_s)$ and $(x_e,y_e)$, its orientation is defined as
\begin{equation}
\theta = \arctan2(y_e - y_s,\, x_e - x_s).
\end{equation}
Let $\{\theta_i\}_{i=1}^{N}$ denote the set of orientations across all skeleton segments within a blob.
Orientation coherence is quantified using the mean resultant length:
\begin{equation}
R = \left| \frac{1}{N} \sum_{i=1}^{N} e^{\mathrm{i}\theta_i} \right|,
\end{equation}
where $R \in [0,1]$.
Higher values indicate stronger orientation alignment (shape-like structure), whereas lower values correspond to more heterogeneous, texture-like edge orientations.

\paragraph{Edge tortuosity.}
To characterize boundary complexity, we measure the tortuosity of skeleton segments within each blob.
For a segment consisting of a sequence of ordered pixels $\{(x_j,y_j)\}_{j=1}^{L}$, tortuosity is defined as the ratio between its path length and the Euclidean distance between endpoints:
\begin{equation}
T = \frac{L}{\sqrt{(x_L-x_1)^2 + (y_L-y_1)^2} + \epsilon},
\end{equation}
where $\epsilon$ is a small constant to ensure numerical stability.
The blob-level tortuosity is obtained by averaging $T$ across all valid skeleton segments.
Higher tortuosity indicates more curved or irregular contours, typically associated with texture-like patterns.

\subsection{Shape-Texture Index (STI)}

To characterize the visual features captured by informative RF blobs, we define a Shape-Texture Index (STI) that jointly accounts for contour organization and boundary complexity.

\paragraph{Texture strength from edge tortuosity.}
Let $T \geq 1$ denote the average tortuosity of skeletonized edge segments within a blob.
To map tortuosity to a bounded texture-related measure, we define the texture strength as
\begin{equation}
\mathrm{Tex}(T) = 1 - \exp\!\big(-\alpha (T - 1)\big),
\end{equation}
where $\alpha > 0$ controls the sensitivity to increasing tortuosity.
This transformation ensures $\mathrm{Tex} \in [0,1)$, with low values corresponding to smooth, contour-like edges and higher values reflecting increasingly irregular, texture-like boundaries.

\paragraph{Shape-Texture Index.}
Let $R \in [0,1]$ denote the orientation coherence of edges within the blob.
The Shape-Texture Index is defined as
\begin{equation}
\mathrm{STI} = R - \mathrm{Tex}(T).
\end{equation}
By construction, $\mathrm{STI} \in (-1,1)$.
Positive values indicate shape-dominated regions characterized by coherent orientation and low tortuosity, whereas negative values correspond to texture-dominated regions with fragmented orientations and complex boundaries.
Values near zero reflect regions that integrate both shape and texture information.

Importantly, the STI does not impose a binary distinction between shape and texture.
Instead, it provides a continuous measure that captures intermediate regimes in which both contour organization and local texture contribute to the representation.

\section{Additional Results}

\paragraph{Similarity Matrices between neural and image feature maps.} 
We compute  pairwise cosine similarity  between image-tower and neural-tower feature maps for all test stimuli.
As shown in Fig.~A.1, similarity matrices exhibit a clear diagonal structure across all four mice, indicating that corresponding image-neural pairs consistently achieve higher similarity than non-matching pairs.

\begin{figure}[t]
    \centering
    \includegraphics[width=\columnwidth]{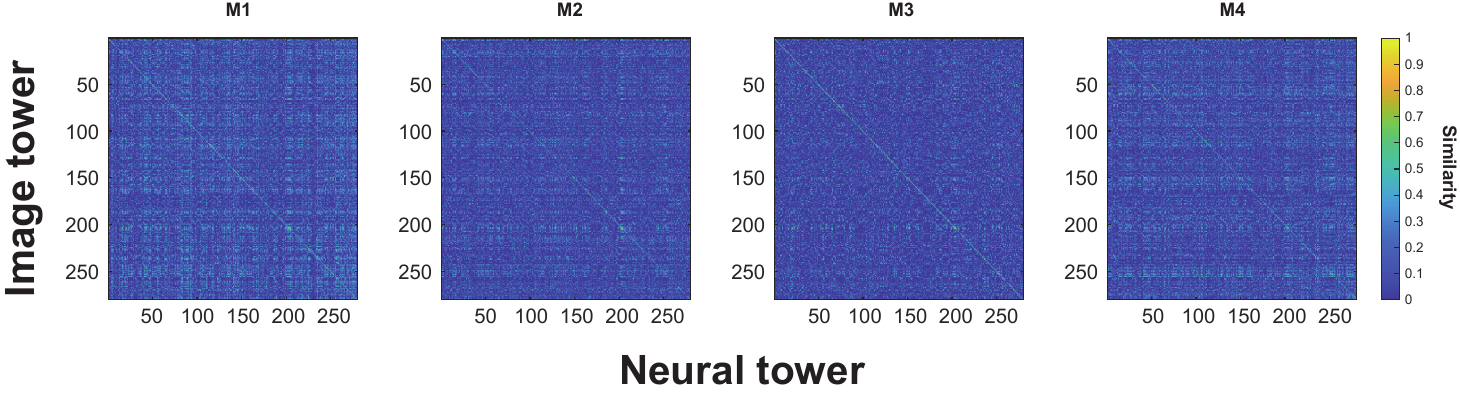}
    \vspace{-6pt}
    \caption{
    \textbf{Image-neural similarity matrices on the test set.}
    For each mouse (M1-M4), we show the pairwise cosine similarity between image-tower feature maps (rows) and neural-tower feature maps (columns) for held-out test stimuli.
    Higher similarity along the diagonal indicates strong correspondence between matched image-neural pairs.
    }
    \label{fig:appendix_similarity_matrix}
\end{figure}

\paragraph{Comparison of aligned feature maps.} 
We compare latent feature maps produced by different baseline architectures, including VGG-MLP and ViT-MLP, with those generated by the proposed dual-pathway model. Although VGG-based models achieve relatively high retrieval accuracy, their aligned feature maps lack coherent spatial organization and exhibit noise-like patterns in both image and neural towers. In contrast, feature maps produced by the proposed model show structured spatial patterns and stronger correspondence between image and neural representations. Additional qualitative examples are provided in Appendix Fig.~A.1.
\begin{figure}[t]
    \centering
    \includegraphics[width=\columnwidth]{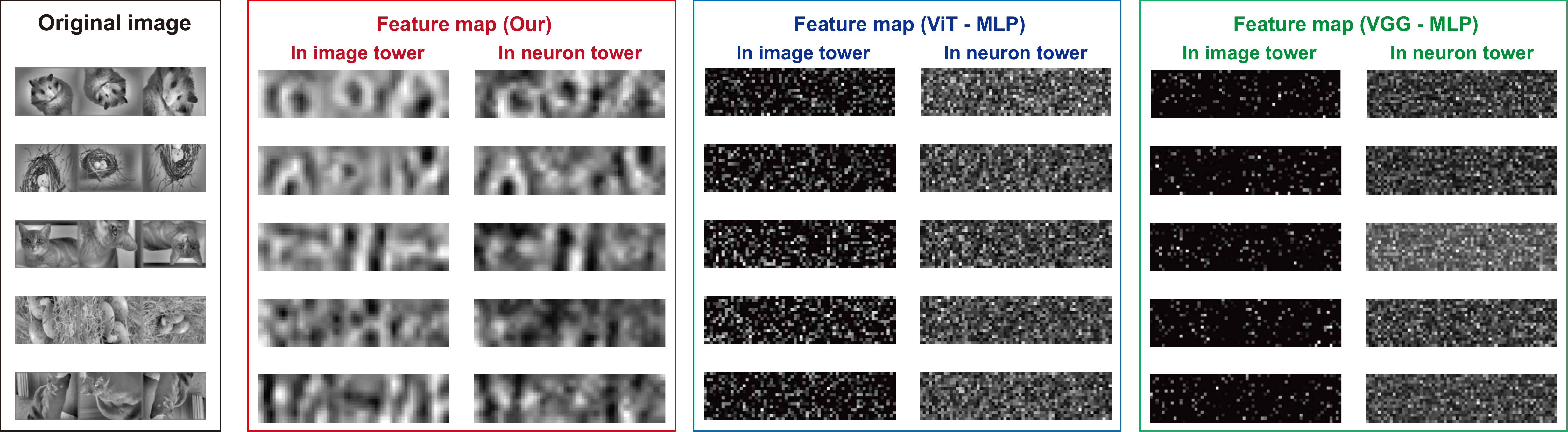}

    \caption{
    \textbf{Qualitative comparison of aligned feature maps across models.}
    Representative original images and the corresponding feature maps from the image and neural towers are shown for our model, ViT-MLP, and VGG-MLP.
    Our model produces feature maps with coherent spatial structure and strong image-neural correspondence, whereas baseline models yield noisy and weakly structured representations.
    }
    \label{fig:appendix_featuremap_comparison}
\end{figure}

\end{document}